\documentclass{article}
\usepackage{spconf,amsmath,graphicx}
\usepackage{algorithm}
\usepackage[]{graphicx}
\usepackage{stfloats}
\usepackage{cite}
\usepackage{psfrag}
\usepackage{wrapfig}
\usepackage{epsfig}
\usepackage{url}
\usepackage{amsmath, amssymb, bm}
\usepackage{array}
\usepackage{times}
\usepackage{multicol}
\usepackage{algorithm}
\usepackage{algorithmic}
\usepackage{mathrsfs}
\usepackage{multirow}
\usepackage{txfonts}
\usepackage[normalem]{ulem}

\usepackage[utf8x]{inputenc}
\usepackage{ucs}
\usepackage{makeidx}
\usepackage[dvipsnames,usenames]{color}
\usepackage{array}
\usepackage{colortbl}
\usepackage{booktabs}
\usepackage{color}
\usepackage{cleveref}
\usepackage{setspace}
\usepackage{subcaption}
\usepackage{enumitem} % remove tab for bullet points during itemize
\usepackage{soul} % for underlining, overstriking, highlighting text
\usepackage{todonotes} % make in-line pdf annotations
\usepackage{balance}
\renewcommand{\vec}[1]{\mathbf{#1}}
\newcommand{\setsym}[1]{\mathbb{#1}}

\graphicspath{ {./fig/} }

\title{Advancing Connectionist Temporal Classification With Attention Modeling}
\name{Amit Das\sthanks{Work performed during an internship at Microsoft.}, Jinyu Li, Rui Zhao, Yifan Gong}
\address{Microsoft AI and Research, One Microsoft Way, Redmond, WA 98052\\
\small \texttt{amitdas@illinois.edu}, \texttt{\{jinyli, ruzhao, ygong\}@microsoft.com}}
\begin{document}
\ninept
\setlength{\abovedisplayskip}{8pt} % set spacing between above text and below eqn
\setlength{\belowdisplayskip}{8pt} % set spacing between above eqn and below text
\maketitle
\begin{abstract}
In this study, we propose advancing all-neural speech recognition by directly incorporating attention modeling within the Connectionist Temporal Classification (CTC) framework. In particular, we derive new context vectors using time convolution features to model attention as part of the CTC network. To further improve attention modeling, we utilize content information extracted from a network representing an implicit language model. Finally, we introduce vector based attention weights that are applied on context vectors  across both time and their individual components. We evaluate our system on a 3400 hours Microsoft Cortana voice assistant task and demonstrate that our proposed model consistently outperforms the baseline model achieving about 20\% relative reduction in word error rates.
\end{abstract}
\begin{keywords}
end-to-end training, CTC, attention, acoustic modeling, speech recognition
\end{keywords}
\vspace{-4mm}
\section{Introduction}
\label{sec: Introduction}\vspace{-2mm}
%Recent advances in automatic speech recognition (ASR) have been mostly due to the advent of deep learning algorithms like Deep Neural Networks (DNNs), Convolutional Neural Networks (CNN), and Recurrent Neural Networks (RNNs). A survey of these algorithms with a focus acoustic modeling is present in \cite{Yu-RecentProgDeepLearningAcousticModels}.  However, one problem with these systems is that there are several intermediate stages (lexicon, language model, decision trees, alignment generation) which either need expert linguistic knowledge or need to be trained separately. 
In the last few years, an emerging trend in automatic speech recognition (ASR) research is the study of end-to-end (E2E) systems \cite{Yu-RecentProgDeepLearningAcousticModels, sak2015learning, miao2015eesen, Chan-LAS, prabhavalkar2017comparison, battenberg2017exploring, sak2017recurrent, hadiantowards, chiu2017state, sainath2017improving}. An E2E ASR system directly transduces an input sequence of acoustic features $\vec{x}$ to an output sequence of probabilities of tokens (phonemes, characters, words etc) $\vec{y}$. This reconciles well with the notion that ASR is inherently a sequence-to-sequence task mapping input  waveforms to output token sequences. Some widely used contemporary E2E approaches for sequence-to-sequence transduction are: (a) Connectionist Temporal Classification (CTC) \cite{Graves-CTCFirst, Graves-E2EASR}, (b) Recurrent Neural Network Encoder-Decoder (RNN-ED) \cite{Cho-RNNEncDecSMT, Bahdanau-RNNEncDecAlignTranslate, Bahdanau-AttentionASR, Chorowski-AttentionASR}, and (c) RNN Transducer \cite{Graves-RNNSeqTransduction}. These approaches have been successfully applied to large scale ASR \cite{sak2015learning, miao2015eesen, Chan-LAS, soltau2016neural, prabhavalkar2017comparison, battenberg2017exploring, rao2017exploring, chiu2017state}.
%An empirical comparison between these approaches can be found in \cite{Hannun-DeepSpeech}. 
%A comprehensive survey of recent advances in ASR, including E2E approaches, is in \cite{Yu-RecentProgDeepLearningAcousticModels}. 
%In this work, we mainly focus on CTC and RNN-ED.

Among the three aforementioned E2E methods, CTC enjoys its training simplicity and is one of the most popular methods used in the speech community (e.g., \cite{Hannun-DeepSpeech, sak2015learning, sak2015fast, miao2015eesen, soltau2016neural, kanda2016maximum, Zweig-AdvancesNeuralASR, liu2017gram, audhkhasi2017direct, Li17CTCnoOOV}). To deal with the issue that the length of the output labels is shorter than the length of the input speech frames, CTC introduces a special blank label and allows for repetition of labels to force the output and input sequences to have the same length. 
%The CTC error criterion tends to improve the traditional cross-entropy (CE) trained system when using context dependent phones as output units and language models (LMs) for decoding \cite{sak2015learning, sak2015fast}.
CTC outputs are usually dominated by blank symbols. The outputs corresponding to the non-blank symbols usually occur with spikes in their posteriors. Thus, an easy way to generate ASR outputs in a CTC network is to concatenate the tokens corresponding to the posterior spikes and collapse them into word outputs if needed. Known as greedy decoding, this is a very attractive feature as there is no involvement of either a language model (LM) or any complex decoding. In \cite{sak2015fast}, a CTC network with up to 27 thousand (27k) word outputs was explored. However, its ASR accuracy is not very appealing, partially due to the high out-of-vocabulary (OOV) rate as a result of using only around 3k hours of training data. Later, in \cite{soltau2016neural}, it was shown that by using 100k words as output targets and by training the model with 125k hours of data, a word-based CTC system, a.k.a. acoustic-to-word CTC, could outperform a phoneme-based CTC system. However, accessibility to more than 100k hours of data is rare. Usually, at most a few thousand hours of data are easily accessible. 

A character-based CTC, with significantly fewer targets, can naturally solve the OOV issue as the word output sequence is generated by collapsing the character output sequence. Because there is no constraint when generating the character output sequence, a character-based CTC in theory can generate any word. However, this is also a drawback of the character-based CTC because it can generate any ill-formed word. As a result, a character-based CTC without any LM and complex decoding usually results in very high word error rates (WER). For example, Google reported around 50\% WER with a character-based CTC on voice search tasks even with 12.5k hours of training data \cite{prabhavalkar2017comparison}. 

The inferior performance of a CTC system stems from two modeling issues. First, CTC relies only on the hidden feature vector at the current time to make predictions. This is the hard alignment problem. Second, CTC imposes the conditional independence constraint that output predictions are independent given the entire input sequence which is not true for ASR. When using larger units such as words as output units, these two issues can be circumvented to some extent. However, when using smaller units such as characters, these two issues become prominent. 

In this study, we focus on reducing the WER of the character-based CTC by addressing its modeling issues. %Our model is still trained with the CTC criterion as we want to retain its training and testing simplicity. 
This is very important to us as we have shown that the OOV issue in a word-based CTC can be solved by using a shared hidden layer hybrid CTC with both words and characters as output units \cite{Li17CTCnoOOV}. Therefore, a high accuracy character-based CTC will benefit our hybrid CTC work in \cite{Li17CTCnoOOV}.

%While both CTC and RNN-ED model the posterior distribution $p(\vec{y}|\vec{x})$, there are some key differences. (a) Attention models using RNN-ED \cite{Bahdanau-AttentionASR, Chan-LAS} have been known to improve ASR performance because of their ability to weigh specific parts of the utterance differently (soft alignment) that are most important to the current prediction. However, RNN-ED does not impose this constraint and can learn language models implicitly.

%A two-stage iterative CTC was proposed in \cite{Zweig-AdvancesNeuralASR} where the noisy output sequence of the first stage CTC is input to a second stage CTC to produce a cleaner output sequence. 

The basic idea for improving CTC is to blend some concepts from RNN-ED into CTC modeling. In the past, several attempts have been made to improve RNN-ED by using CTC as an auxiliary task in a multitask learning (MTL) framework, either at the top layer \cite{Kim-JointCTCRNNEncDecUsingMTL, hori2017advances} or at the intermediate encoder layer \cite{Toshniwal-MTLLowLevelRNNED}.
Our motivation in this work is to address the hard alignment problem of CTC, as outlined earlier, by modeling attention directly within the CTC framework. This differs from previous approaches where attention and CTC models were part of separate sub-networks in an MTL framework. To this end, we propose the following key ideas in this paper. (a) First, we derive context vectors using \textit{time convolution features} (Sec \ref{ssec: CTCAttn-conv}) and apply attention weights on these context vectors (Sec \ref{ssec: CTCAttn-attn}). This makes it possible for CTC to be trained using soft alignments instead of hard. (b) Second, to improve attention modeling, we incorporate an \textit{implicit language model} (Sec \ref{ssec: CTCAttn-LM}) during CTC training. (c) Finally, we extend our attention modeling further by introducing \textit{component attention} (Sec \ref{ssec: CTCAttn-Comp}) where context vectors are weighted across both time and their individual components. %The remainder of this paper is organized as follows. In Section \ref{sec: E2E}, we present a brief overview of CTC and RNN-ED. We explain CTC with attention in Section \ref{sec: CTCAttn} followed by experiments and results in Section \ref{sec: ExptsResults}.

\vspace{-8mm}
\section{End-to-End Speech Recognition}\vspace{-2mm}
\label{sec: E2E}
An E2E ASR system models the posterior distribution $p(\vec{y}|\vec{x})$ by transducing an input sequence of acoustic feature vectors to an output sequence of vectors of posterior probabilities of tokens. More specifically, for an input sequence of feature vectors $\vec{x} = (\vec{x}_{1}, \cdots, \vec{x}_{T})$ of length $T$ with $\vec{x}_{t} \in \setsym{R}^{m}$, an E2E ASR system transduces the input sequence to an intermediate sequence of hidden feature vectors $\vec{h} = (\vec{h}_{1}, \cdots, \vec{h}_{L})$ of length $L$ with $\vec{h}_{l} \in \setsym{R}^{n}$. The sequence $\vec{h}$ undergoes another transduction resulting in an output sequence of vectors of posterior probabilities $p(\vec{y}|\vec{x})$ where $\vec{y} = (\vec{y}_{1}, \cdots, \vec{y}_{U})$ of length $U$ with $\vec{y}_{u} \in \setsym{L}^{K}$, $\setsym{L}$ being the label set, such that $\sum_{k=1}^{K} p(y_{u}(k)|\vec{x}) = 1$. Here, $K = \left\vert\setsym{L}\right\vert$ is the cardinality of the label set $\setsym{L}$. Usually $U \leq T$ in E2E systems. %In addition, $y_{u}(k)$ is the $k^{th}$ element of $\vec{y}_{u}$ and represents the probability of label $k$ at time $u$ such that $\sum_{k=1}^{K} y_{u}(k) = 1$. In ASR, labels could be senones, graphemes, letters, words etc. depending on the desired granularity of outputs. Usually $U \leq T$ since the length of the output sequence is smaller than the length of the input sequence.
%The E2E system models the joint posterior distribution $p(\vec{y}|\vec{x})$ using neural networks trained using large amounts of sequential data $\vec{x}$ and corresponding labels $\vec{y}$. Thus, a neural network, parameterized by $\mathbf{W}$, learns a many-to-one functional $\vec{f}_{\mathbf{W}}: \vec{x} \mapsto \vec{y}$ which closely resembles $p(\vec{y}|\vec{x})$.

\vspace{-4mm}
\subsection{Connectionist Temporal Classification (CTC)}\vspace{-2mm}
\label{ssec: CTC}
A CTC network uses an RNN and the CTC error criterion \cite{Graves-CTCFirst, Graves-E2EASR}  which directly optimizes the prediction of a transcription sequence. Since RNN operates at the frame level, the lengths of the input sequence $\vec{x}$ and the output sequence $\vec{y}$ must be the same, i.e., $T = L = U$. To achieve this, CTC adds a \textit{blank} symbol as an additional label to the label set $\setsym{L}$ and allows repetition of labels or blank across frames. Since the RNN generates a lattice of posteriors $p(\vec{y}|\vec{x})$, it undergoes additional post-processing steps to produce another human readable sequence (a transcription). More specifically, each path through the lattice $p(\vec{y}|\vec{x})$ is a sequence of labels at the frame level and is known as the CTC path. Denoting any CTC path as $\bm\pi$ of length $T$ and the label of that path at time $t$ as $\pi_{t}$, we have the path probability of $\bm\pi$ as,
\vspace{-2mm}
\begin{align}
p(\bm\pi|\vec{x}) &\overset{\text{CI}}{=} \prod_{t=1}^{T} p(\pi_{t}|\vec{x}), \label{eq:ctcpathprob}
\end{align}
where the equality is based on the conditional independence assumption, i.e., $(\pi_{t} \Perp \pi_{\ne t})|\vec{x}$. %that the network output at time $t$ is conditionally independent of outputs at other times given $\vec{x}$, i.e. $(\pi_{t} \Perp \pi_{\ne t})|\vec{x}$. %$y_{t}(\pi_{t}) = p(\pi_{t}|\vec{x})$, is conditionally independent of outputs at other times given $\vec{x}$, i.e. $p(\pi_{\neq t}|\vec{x})$.
Due to this constraint, CTC does not model inter-label dependencies. Therefore, during decoding it relies on external language models to achieve good ASR accuracy. More details about CTC training are covered in \cite{Graves-CTCFirst, Graves-E2EASR}.

\vspace{-4mm}
\subsection{RNN Encoder-Decoder (RNN-ED)}\vspace{-2mm}
\label{ssec: RNNEncDec}
An RNN-ED \cite{Cho-RNNEncDecSMT, Bahdanau-RNNEncDecAlignTranslate, Bahdanau-AttentionASR, Chorowski-AttentionASR} uses two distinct networks - an RNN encoder network that transforms $\vec{x}$ into $\vec{h}$ and an RNN decoder network that transforms $\vec{h}$ into $\vec{y}$. Using these, an RNN-ED models $p(\vec{y}|\vec{x})$ as,
\vspace{-2mm}
\begin{align}
p(\vec{y}|\vec{x}) &= \prod_{u=1}^{U} p(\vec{y}_{u}|\vec{y}_{1:u-1}, \vec{c}_{u}), \label{eq:RNNED-transcriptprob}
\end{align}
where $\vec{c}_{u}$ is the context vector at time $u$ and is a function of $\vec{x}$.
There are two key differences between CTC and RNN-ED. First, $p(\vec{y}|\vec{x})$ in \eqref{eq:RNNED-transcriptprob} is generated using a product of ordered conditionals. Thus, RNN-ED is not impeded by the conditional independence constraint of \eqref{eq:ctcpathprob}. Second, the lengths of the input and output sequences are allowed to differ (i.e., $T = L > U$). The decoder output $\vec{y}_{u}$ at time $u$ is dependent on $\vec{c}_{u}$ which is a weighted sum of all its inputs, i.e., $\vec{h}_{t}, t = 1, \cdots, T$. On the contrary, CTC generates $\vec{y}_{t}$ using only $\vec{h}_{t}$.
%More specifically, the encoder network computes,
%\vspace{-2mm}
%\begin{align}
%\vec{h}_{t} &= \text{Encode}(\vec{x}_{t}, \vec{h}_{t-1}). \label{eq:RNNED-encode}
%\end{align}

The decoder network of RNN-ED has three components: a multinomial distribution generator \eqref{eq:RNNED-generate}, an RNN decoder \eqref{eq:RNNED-recurrent}, and an attention network \eqref{eq:RNNED-annotate}-\eqref{eq:RNNED-locfeat} as follows:
\begin{align}
p(\vec{y}_{u}|\vec{y}_{1:u-1}, \vec{c}_{u}) &= \text{Generate}(\vec{y}_{u-1}, \vec{s}_{u}, \vec{c}_{u}), \label{eq:RNNED-generate} \\
\vec{s}_{u} &= \text{Recurrent}(\vec{s}_{u-1}, \vec{y}_{u-1}, \vec{c}_{u}), \label{eq:RNNED-recurrent} \\
\vec{c}_{u} &= \text{Annotate}(\bm\alpha_{u}, \vec{h}) = \sum_{t=1}^{T} \alpha_{u,t} \vec{h}_{t}, \label{eq:RNNED-annotate} \\
\bm\alpha_{u} &= \text{Attend}(\vec{s}_{u-1}, \bm\alpha_{u-1}, \vec{h}). \label{eq:RNNED-attend}
\end{align}
Here, $\vec{h}_{t}, \vec{c}_{u} \in \setsym{R}^{n}$ and $\bm\alpha_{u} \in \setsym{U}^{T}$, where $\setsym{U} = [0, 1]$,  such that $\sum_t \alpha_{u,t} = 1$. Also, for simplicity assume $\vec{s}_{u} \in \setsym{R}^{n}$. $\text{Generate}(.)$ is a feedforward network with a softmax operation generating the ordered conditional $p(\vec{y}_{u}|\vec{y}_{1:u-1}, \vec{c}_{u})$ \vphantom{for my reference only,see \cite[Appendix A.2.2]{Bahdanau-RNNEncDecAlignTranslate}}.  Recurrent(.) is an RNN decoder operating on the output time axis indexed by $u$ and has hidden state $\vec{s}_{u}$. Annotate(.) computes the context vector $\vec{c}_{u}$ (also called the soft alignment) using the attention probability vector $\bm\alpha_{u}$ and the hidden sequence $\vec{h}$. Attend(.) computes the attention weight $\alpha_{u,t}$ using a single layer feedforward network as,
\begin{align}
e_{u,t} &= \text{Score}(\vec{s}_{u-1}, \bm\alpha_{u-1}, \vec{h}_{t}), \quad t = 1, \cdots, T \label{eq:RNNED-score} \\
\alpha_{u, t} &= \frac{ \text{exp}(e_{u, t}) } { \sum_{t^{\prime}=1}^{T} \text{exp}(e_{u, t^{\prime}}) }, \label{eq:RNNED-normalizedscore}
\end{align}
where $e_{u, t} \in \setsym{R}$. $\text{Score}(.)$ can either be content-based attention or hybrid-based attention. The latter encodes both content ($\vec{s}_{u-1}$) and location ($\bm\alpha_{u-1}$) information. $\text{Score}(.)$ is computed using,
\begin{align}
\hspace{-2mm} e_{u, t} &= \begin{cases}
\vec{v}^{T}\text{tanh}\ (\vec{U} \vec{s}_{u-1} + \vec{W} \vec{h}_{t}  +  \vec{b}), \ \mbox{(content)} \\
\vec{v}^{T}\text{tanh}\ (\vec{U} \vec{s}_{u-1} + \vec{W} \vec{h}_{t}  + \vec{V} \vec{f}_{u,t} + \vec{b}), \ \mbox{(hybrid)}
\end{cases} \label{eq:RNNED-ContentHybrid} \\
&\text{where,} \quad \vec{f}_{u,t} = \vec{F} \ast \bm\alpha_{u-1} \label{eq:RNNED-locfeat}.
\end{align}
The operation $\ast$ denotes convolution. Attention parameters $\vec{U}, \vec{W}, \vec{V}$, $\vec{F}, \vec{b}, \vec{v}$ are learned while training RNN-ED.

\vspace{-4mm}
\section{CTC With Attention}
\begin{figure*}
\centering
\resizebox{0.90\linewidth}{!}{
\includegraphics[width=\textwidth,height=0.48\textwidth,trim=4mm 0mm 2mm 6mm,clip]{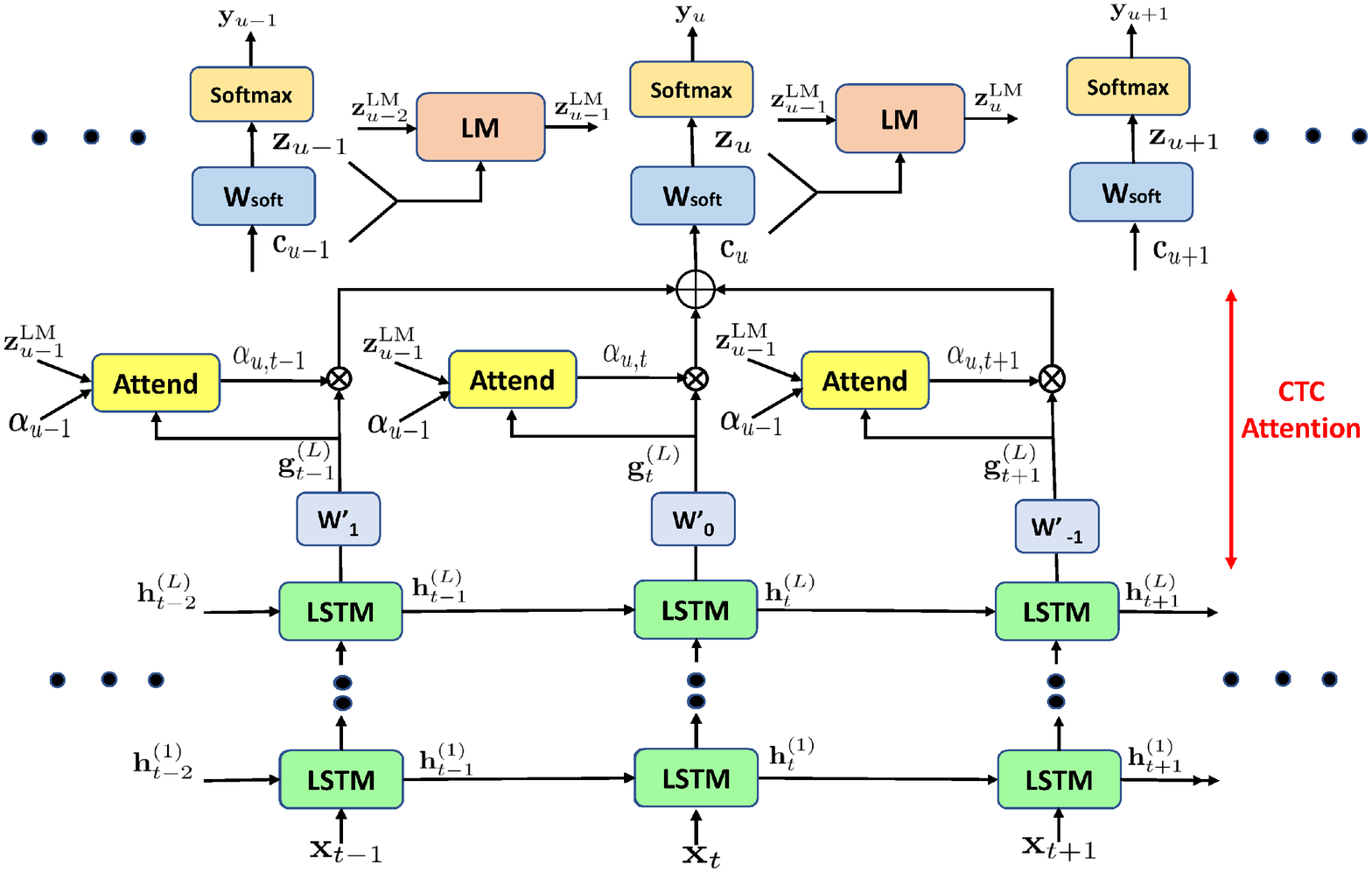}\vspace{-2mm}
}
\caption{An example of a CTC Attention network with an attention window of size $C = 3 \ (\text{i.e., } \tau = 1)$.}
\label{fig:CTCAttn}\vspace{-4mm}
\end{figure*}

\vspace{-2mm}
\label{sec: CTCAttn}
In this section, we outline various steps required to model attention directly within CTC. An example of the proposed CTC attention network is shown in Figure \ref{fig:CTCAttn}.
Since our network is basically a CTC network, the input and output sequences are of the same length (i.e., $T = U$). However, we will use the indices $t$ and $u$ to denote the time step for input and output sequences respectively. This is only to maintain notational consistency with the equations in RNN-ED. It is understood that every input frame $\vec{x}_{t}$ generates output $\vec{y}_{t} = \vec{y}_{u}$.

\vspace{-4mm}
\subsection{Time Convolution (TC) Features}\vspace{-2mm}
\label{ssec: CTCAttn-conv}
Consider a rank-3 tensor $\vec{W}^{\prime} \in \setsym{R}^{n_{1} \times n_{2} \times C}$. For simplicity, assume $n_{1} = n_{2} = n$ where $n$ is the dimension of the hidden feature $\vec{h}_{t}$. Our attention model considers a small subsequence of $\vec{h}$ rather than the entire sequence. This subsequence, $(\vec{h}_{u-\tau}, \cdots, \vec{h}_{u}, \cdots, \vec{h}_{u+\tau})$, will be referred to as the attention window. Its length is $C$ and it is centered around the current time $u$.  Let $\tau$ represent the length of the window on either side of $u$. Thus, $C = 2\tau + 1$. Then $\vec{c}_{u}$ can be computed using,
\begin{align}
\vec{c}_{u} &= \vec{W}^{\prime} \ast \vec{h} = \sum_{t =u-\tau}^{u+\tau} \vec{W}^{\prime}_{u - t} \vec{h}_{t} \nonumber \\
			&\stackrel{\Delta}{=} \sum_{t =u-\tau}^{u+\tau} \vec{g}_{t} = \gamma \sum_{t =u-\tau}^{u+\tau} \alpha_{u,t} \vec{g}_{t}. \label{eq:CTCAttn-TimeConvolution}
\end{align}
Here, $\vec{g}_{t} \in \setsym{R}^{n}$ represents the $filtered$ signal at time $t$. The last step \eqref{eq:CTCAttn-TimeConvolution} holds when $\alpha_{u,t} = \frac{1}{C}$ and $\gamma = C$. Since \eqref{eq:CTCAttn-TimeConvolution}
is similar to \eqref{eq:RNNED-annotate} in structure, $\vec{c}_{u}$ represents a special case context vector with \textit{uniform} attention weights $\alpha_{u,t} = \frac{1}{C}$, $t \in [u-\tau, \ u+\tau]$. Also, $\vec{c}_{u}$ is a result of convolving features $\vec{h}$ with $\vec{W}^{\prime}$ in time. Thus, $\vec{c}_{u}$ represents a time convolution feature. This is illustrated in Figure \ref{fig:CTCAttn} for the case of $\tau = 1$ (after ignoring the Attend block and letting $\alpha_{u,t} = \frac{1}{C}$).

\vspace{-4mm}
\subsection{Content Attention (CA) and Hybrid Attention (HA)}\vspace{-2mm}
\label{ssec: CTCAttn-attn}
To incorporate non-uniform attention in \eqref{eq:CTCAttn-TimeConvolution}, we need to compute a non-uniform $\alpha_{u, t}$ for each $t \in [u-\tau, \ u+\tau]$ using the attention network in \eqref{eq:RNNED-attend}. %Here, $\bm\alpha_{u} \in \setsym{U}^{C}$ instead of $\bm\alpha_{u} \in \setsym{U}^{T}$ in \eqref{eq:RNNED-annotate}.
However, since there is no explicit decoder like \eqref{eq:RNNED-recurrent} in CTC, there is no decoder state $\vec{s}_{u}$. Therefore, we use $\vec{z}_{u}$ instead of $\vec{s}_{u}$. The term $\vec{z}_{u}  \in \setsym{R}^{K}$ is the logit to the softmax and is given by,
\vspace{-2mm}
\begin{align}
\vec{z}_{u} &= \vec{W}_{\text{soft}}\vec{c}_{u} + \vec{b}_{\text{soft}}, \nonumber \\
\vec{y}_{u} &= \text{Softmax}(\vec{z}_{u}), \label{eq:CTCAttn-generate}
\end{align}
where  $\vec{W}_{\text{soft}} \in \setsym{R}^{K \times n}, \vec{b}_{\text{soft}} \in \setsym{R}^{K}$. Thus, \eqref{eq:CTCAttn-generate} is similar to the Generate(.) function in \eqref{eq:RNNED-generate} but lacks the dependency on $\vec{y}_{u-1}$ and $\vec{s}_{u}$. Consequently, the Attend(.) function in \eqref{eq:RNNED-attend} becomes,
\begin{align}
\bm\alpha_{u} &= \text{Attend}(\vec{z}_{u-1}, \bm{\alpha}_{u-1}, \vec{g}), \label{eq:CTCAttn-attend}
\end{align}
where $\vec{h}$ in $\eqref{eq:RNNED-attend}$ is replaced with $\vec{g} = (\vec{g}_{u - \tau}, \cdots, \vec{g}_{u + \tau})$.
%Note that it is possible to substitute $\vec{z}_{u}$ with  $\vec{y}_{u}$ because of the one-to-one mapping of the softmax function.
Next, the scoring function Score(.) in \eqref{eq:RNNED-score} is modified by replacing the raw signal $\vec{h}_{t}$ with the filtered signal $\vec{g}_{t}$. Thus, the new Score(.) function becomes,
\vspace{-4mm}
\begin{align}
e_{u,t} &= \text{Score}(\vec{z}_{u-1}, \bm\alpha_{u-1}, \vec{g}_{t}), \label{eq:CTCAttn-score} \\
&= \begin{cases}
\vec{v}^{T}\text{tanh}(\vec{U} \vec{z}_{u-1} + \vec{W} \vec{g}_{t} + \vec{b}), \ \mbox{(content)} \\
\vec{v}^{T}\text{tanh}(\vec{U} \vec{z}_{u-1} + \vec{W} \vec{g}_{t} + \vec{V} \vec{f}_{u,t} + \vec{b}) \ \mbox{(hybrid)}
\end{cases} \label{eq:CTCAttn-ContentHybrid}
\end{align}
with $\vec{f}_{u,t}$ a function of $\bm\alpha_{u-1}$ through \eqref{eq:RNNED-locfeat}. The content and location information are encoded in $\vec{z}_{u-1}$ and $\bm\alpha_{u-1}$ respectively. The role of $\vec{W}$ in \eqref{eq:CTCAttn-ContentHybrid} is to project $\vec{g}_{t}$ for each $t \in [u-\tau, \ u+\tau]$ to a common subspace. Score normalization of \eqref{eq:CTCAttn-score} can be achieved using \eqref{eq:RNNED-normalizedscore} to generate non-uniform $\alpha_{u, t}$ for $t \in [u-\tau, \ u+\tau]$. Now, $\bm\alpha_{u}$ can be plugged into \eqref{eq:CTCAttn-TimeConvolution}, along with $\vec{g}$ to generate the context vector $\vec{c}_{u}$. This completes the attention network. We found that excluding the scale factor $\gamma$ in \eqref{eq:CTCAttn-TimeConvolution}, even for non-uniform attention, was detrimental to the final performance. Thus, we continue to use $\gamma = C$.

\vspace{-4mm}
\subsection{Implicit Language Model (LM)}\vspace{-2mm}
\label{ssec: CTCAttn-LM} 
The performance of the attention model can be improved further by providing more reliable content information from the previous time step. This is possible by introducing another recurrent network that can utilize content from several time steps in the past. This network, in essence, would learn an implicit LM. In particular, we feed $\vec{z}^{\text{LM}}_{u-1}$ (hidden state of the LM network) instead of $\vec{z}_{u-1}$ to the Attend(.) function in \eqref{eq:CTCAttn-attend}. To build the LM network, we follow an architecture similar to RNN-LM \cite{Mikolov-RNNLM}. As illustrated in the LM block of Figure \ref{fig:CTCAttn}, the input to the network is computed by stacking the previous output $\vec{z}_{u-1}$ with the context vector $\vec{c}_{u-1}$ and feeding it to a recurrent function $\mathcal{H}(.)$. This is represented as,
\begin{align}
\vec{z}^{\text{LM}}_{u-1} &= \mathcal{H}(\vec{x}_{u-1}, \vec{z}^{\text{LM}}_{u-2}), \quad
\vec{x}_{u-1} = 
\begin{bmatrix}
\vec{z}_{u-1} \\
\vec{c}_{u-1}
\end{bmatrix}, \label{eq:CTCAttnLM-LSTM} \\
\bm\alpha_{u} &= \text{Attend}(\vec{z}^{\text{LM}}_{u-1}, \bm{\alpha}_{u-1}, \vec{g}).  \label{eq:CTCAttn-attendLM}
\end{align}
We model $\mathcal{H}(.)$ using a long short-term memory (LSTM) unit \cite{Hochreiter1997long} with $n$ memory cells and input and output dimensions set to $K + n$ ($\vec{x}_{u-1} \in \setsym{R}^{K+n}$) and $n$ ($\vec{z}^{\text{LM}}_{u-1} \in \setsym{R}^{n}$) respectively. One problem with $\vec{z}^{\text{LM}}_{u-1}$ is that it encodes the content of a pseudo LM, rather than a true LM, since CTC outputs are interspersed with blank symbols by design. Another problem is that $\vec{z}^{\text{LM}}_{u-1}$ is a real-valued vector instead of a one-hot vector. Hence, this LM is an implicit LM rather than an explicit or a true LM.

\vspace{-4mm}
\subsection{Component Attention (COMA)}\vspace{-2mm}
\label{ssec: CTCAttn-Comp}
In the previous sections, $\alpha_{u,t} \in \setsym{U}$ is a scalar term weighting the contribution of the vector $\vec{g}_{t} \in \setsym{R}^{n}$ to generate the output $\vec{y}_{u}$ through \eqref{eq:CTCAttn-TimeConvolution} and \eqref{eq:CTCAttn-generate} . This means all $n$ components of the vector $\vec{g}_{t}$ are weighted by the same scalar $\alpha_{u,t}$. In this section, we consider weighting each component of $\vec{g}_{t}$ distinctively. Therefore, we need a vector weight $\bm\alpha_{u,t} \in \setsym{U}^{n}$ instead of the scalar weight $\alpha_{u,t} \in \setsym{U}$ for each $t \in [u-\tau, \ u+\tau]$. The vector $\bm\alpha_{u,t}$ is generated by first computing an $n$-dimensional score $\vec{e}_{u, t}$ for each $t$. This is easily achieved using the Score(.) function in \eqref{eq:CTCAttn-ContentHybrid} but without taking the inner product with $\vec{v}$. For example, in the case of hybrid, the scoring function becomes,
\begin{align}
\vec{e}_{u, t} &= \text{tanh}(\vec{U} \vec{z}_{u-1} + \vec{W} \vec{g}_{t}  + \vec{V} \vec{f}_{u,t} + \vec{b}). \label{eq:CTCAttn-Comp-score}
\end{align}
Now, we have $C$ column vectors $[\vec{e}_{u, u-\tau}, \cdots, \vec{e}_{u, u+\tau}]$ where each $\vec{e}_{u, t} \in (-1, 1)^{n}$. Let $e_{u, t, j} \in (-1,1)$ be the $j^{\text{th}}$ component of the vector $\vec{e}_{u, t}$. To compute $\alpha_{u, t, j}$ from $e_{u, t, j}$, we normalize $e_{u, t, j}$ across $t$ keeping $j$ fixed. Thus, $\alpha_{u, t, j}$ is computed as,
\vspace{-2mm}
\begin{align}
\alpha_{u, t, j} &= \frac{\text{exp}(e_{u, t, j})}{\sum_{t^{\prime}=u-\tau}^{u+\tau} \text{exp}(e_{u, t^{\prime}, j})}, \quad j=1,\cdots,n. \label{eq:CTCAttn-Comp-scoresoftmax}
\end{align}
Here, $\alpha_{u, t, j}$ can be interpreted as the amount of contribution from $g_{t}(j)$ in computing $c_{u}(j)$.
Now, from \eqref{eq:CTCAttn-Comp-scoresoftmax}, we know vectors $\bm\alpha_{u,t}$ for each $t \in [u-\tau, \ u+\tau]$.
Under the COMA formulation, the context vector $\vec{c}_{u}$ can be computed using,
\setlength{\abovedisplayskip}{-3pt}
\setlength{\belowdisplayskip}{-3pt}
\begin{align}
\vec{c}_{u} &= \text{Annotate}(\bm\alpha_{u}, \vec{g}, \gamma) = \gamma \sum_{t=u-\tau}^{u+\tau} \bm\alpha_{u,t} \odot \vec{g}_{t}, \label{eq:CTCAttn-Comp-annotate}
\end{align}
where $\odot$ is the Hadamard product.

Note that as extensions of CTC, both RNN-T \cite{Graves-RNNSeqTransduction, rao2017exploring} and RNN aligner \cite{sak2017recurrent} either change the objective function or the training process to relax the frame independence assumption of CTC. The proposed attention CTC is another solution by working on hidden layer representation with more context information without changing the CTC objective function and training process.

\vspace{-4mm}
\section{Experiments}\vspace{-2mm}
\label{sec: Expts}
The proposed methods were evaluated using transcribed data collected from Microsoft's Cortana voice assistant system.  The training set consists of about 3.3 million short utterances ($\sim$ 3400 hours) in US-English. The test set consists of about 5600 utterances ($\sim$ 6 hours). All CTC models were trained on top of either uni-directional or bi-directional 5-layer LSTMs. The uni-directional LSTM has 1024 memory cells while the bi-directional one has 512 memory cells in each direction (therefore still 1024 output dimensions when combining outputs from both directions). Then they are linearly projected to 512 dimensions. The base feature vector computed every 10 ms frame is a 80-dimensional vector containing log filterbank energies. Eight frames of base features were stacked together (hence, $m = 80 \times 8 = 640$) as the input to the uni-directional CTC, while three frames were stacked together ($m = 240$) for the bi-directional CTC. The skip size for both uni- and bi-directional CTC was three frames as in \cite{sak2015fast}. The dimension $n$ of vectors $\vec{h}_{t}, \vec{g}_{t}, \vec{c}_{u}$ was set to 512. Character-based CTC was used in all our experiments. For decoding, we use the greedy decoding procedure (no complex decoder or external LM). Thus, our system is a pure all-neural system. %Furthermore, neither any external LM nor any complex decoder was used at any stage. We directly concatenate the characters with maximum posteriors and collapse them into words. Thus, our system is a pure all-neural system. Instead of character error rate, we used the more challenging word error rate metric for our evaluations.

\vspace{-4mm}
\subsection{Unidirectional CTC with 28-character set}\vspace{-2mm}
\label{ssec: UniCTC_28Char}
In the first set of experiments, the vanilla CTC \cite{Graves-CTCFirst} and the proposed CTC models were evaluated with a unidirectional 5-layer LSTM. The output layer has 28 output nodes (hence, K = 28) corresponding to a 28-character set (26 letters ‘a’-‘z’ + space + blank). $\tau$ was empirically set to 4, which means the context window size ($C$) for attention was 9. The results are tabulated in Table \ref{Tab:WER_UniCTC_28Char}. The top row summarizes the WER for vanilla CTC. All subsequent rows under ``CTC Attention" summarize the WER for the proposed CTC models when attention modeling capabilities were gradually added in a stage-wise fashion. The best CTC Attention model is in the last row and it outperforms the vanilla CTC model by 18.75\% relative. There is a slight increase in WER when adding HA on top of CA. In general, for the other experiments, we find that adding HA is beneficial although the gains are marginal compared to all the other enhancements (CA, LM, COMA). Benefits of location based attention could become more pronounced when attention spans over very large contexts \cite{Bahdanau-AttentionASR}. %However, our experiments with larger contexts did not yield improvements consistently.

\begin{table}[!t]
\centering
\caption{WERs of Vanilla CTC and CTC Attention models for $\tau = 4\ (C = 9)$ %$\tau = 8\ (C = 17)$
trained with unidirectional 5-layer LSTM and 28-character set. Relative WER improvements are in parentheses.}
\vspace{-3mm}
%\begin{tabular}{l|c c}
   %\hline
%E2E Model  & \multicolumn{2}{c}{WER (\%)} \\
         		   						  %&$\tau = 4$      &$\tau = 8$     \\ \hline
%Vanilla CTC 						    &29.60 (0.00)    &29.60 (0.00)   \\ \hline
%CTC Attention      						  &   			   &  			   \\
%TC  (Sec \ref{ssec: CTCAttn-conv})            &27.36 (07.56)   &27.76 (06.22)  \\
%+CA (Sec \ref{ssec: CTCAttn-attn})            &25.41 (14.16)   &24.14 (18.45)  \\
%+HA (Sec \ref{ssec: CTCAttn-attn})            &25.62 (13.45)   &23.94 (19.12)  \\
%+LM (Sec \ref{ssec: CTCAttn-LM})              &24.74 (16.42)   &23.68 (20.00)  \\
%+COMA (Sec \ref{ssec: CTCAttn-Comp})          &\bf{24.05 (18.75)}   &\bf{23.25 (21.45)}  \\ \hline
%\end{tabular}

\begin{tabular}{l|c c}
   \hline
E2E Model  & \multicolumn{1}{c}{WER (\%)} \\
 \hline
Vanilla CTC \cite{Graves-CTCFirst} 	      &29.60 (0.00)       \\ \hline
CTC Attention      						  &   			   \\
TC  (Sec \ref{ssec: CTCAttn-conv})            &27.36 (07.56)     \\
+CA (Sec \ref{ssec: CTCAttn-attn})            &25.41 (14.16)     \\
+HA (Sec \ref{ssec: CTCAttn-attn})            &25.62 (13.45)     \\
+LM (Sec \ref{ssec: CTCAttn-LM})              &24.74 (16.42)     \\
+COMA (Sec \ref{ssec: CTCAttn-Comp})          &\bf{24.05 (18.75)}     \\ \hline
\end{tabular}
\vspace{-3mm}
\label{Tab:WER_UniCTC_28Char}
\end{table}

\vspace{-4mm}
\subsection{Bidirectional CTC with 28-character set}\vspace{-2mm}
\label{ssec: BiCTC_28Char}
In the next set of experiments, the baseline and proposed CTC models were evaluated with a bidirectional 5-layer LSTM with  $\tau = 4$ using the 28-character set. Otherwise, we followed the same training regime as in Section \ref{ssec: UniCTC_28Char}. The results are tabulated in Table \ref{Tab:WER_BiCTC_28Char}. Similar to the unidirectional case, the best CTC Attention model outperforms vanilla CTC by about 21.06\% relative.
This shows that even a strong baseline like bidirectional CTC does not undermine the efficacy of the proposed CTC Attention models.

\begin{table}[!t]
\centering
\caption{WERs of Vanilla CTC and CTC Attention models for $\tau = 4\ (C = 9)$ trained with bidirectional 5-layer LSTM and 28-character set. Relative WER improvements are in parentheses.}
\vspace{-3mm}
\begin{tabular}{l|c}
   \hline
E2E Model  & \multicolumn{1}{c}{WER (\%)} \\
            \hline
Vanilla CTC \cite{Graves-CTCFirst}    &26.36 (0.00)    \\ \hline
CTC Attention      					  &  \\
TC    (Sec \ref{ssec: CTCAttn-conv})      &25.21 (04.36)     \\
+CA   (Sec \ref{ssec: CTCAttn-attn})      &22.73 (13.77)     \\
+HA   (Sec \ref{ssec: CTCAttn-attn})      &22.52 (14.57)     \\
+LM   (Sec \ref{ssec: CTCAttn-LM})        &21.69 (17.72)     \\
+COMA (Sec \ref{ssec: CTCAttn-Comp})	  &\bf{20.81 (21.06)}     \\ \hline
\end{tabular}
\vspace{-3mm}
\label{Tab:WER_BiCTC_28Char}
\end{table}

\vspace{-3mm}
\subsection{Bidirectional CTC with 83-character set}\vspace{-2mm}
\label{ssec: BiCTC_83Char}
In the final set of experiments, in addition to the bidirectional LSTM, we construct a new character set \cite{Zweig-AdvancesNeuralASR} by adding new characters on top of the 28-character set. These additional characters include capital letters used in the word-initial position, double-letter units representing repeated characters like \textit{ll}, apostrophes followed by letters such as \textit{`de}, \textit{`r} etc.  Readers may refer \cite{Zweig-AdvancesNeuralASR} for more details. Altogether such a large unit inventory has 83 characters, and we refer to it as the 83-character set. The results for this experiment are tabulated in Table \ref{Tab:WER_BiCTC_83Char}. Again, CTC Attention models consistently outperform vanilla CTC with the best relative improvement close to 20.61\%. This shows that the proposed CTC attention network can achieve similar improvements, no matter whether the vanilla CTC is built with advanced modeling capabilities (from uni-directional to bi-directional) or different sets of character units (28 vs. 83 units).

\begin{table}[!t]
\centering %\begin{center} does the same thing as \centering but inserts an extra line
\caption{WERs of Vanilla CTC and CTC Attention models for $\tau = 4\ (C = 9)$ trained with bidirectional 5-layer LSTM and 83-character set. Relative WER improvements are in parentheses.}
\vspace{-3mm}
\begin{tabular}{l|c}
   \hline
E2E Model  & \multicolumn{1}{c}{WER (\%)} \\
                  \hline
Vanilla CTC \cite{Graves-CTCFirst}    &23.29 (0.00)  \\ \hline
CTC Attention   &  				  \\
TC    (Sec \ref{ssec: CTCAttn-conv})      &22.30 (04.25)      \\
+CA   (Sec \ref{ssec: CTCAttn-attn})      &21.34 (08.37)      \\
+HA   (Sec \ref{ssec: CTCAttn-attn})      &20.81 (10.65)      \\
+LM   (Sec \ref{ssec: CTCAttn-LM})        &19.98 (14.21)      \\
+COMA (Sec \ref{ssec: CTCAttn-Comp})	  &\bf{18.49 (20.61)} \\ \hline
\end{tabular}
\vspace{-6mm}
\label{Tab:WER_BiCTC_83Char}
\end{table}
  
\vspace{-4mm}
\section{Conclusions}\vspace{-2mm}
\label{sec: Conclusions}
In this study, we proposed advancing CTC by directly incorporating attention modeling into the CTC framework. We accomplished 
this by using time convolution features, non-uniform attention, implicit language modeling, and component attention. Our experiments demonstrated that CTC Attention consistently outperformed vanilla CTC by around 20\% relative improvement in WER, no matter whether the vanilla CTC is built with advanced modeling capabilities or different sets of character units. 
As has been reported in \cite{Li18CTCnoOOV}, the proposed method can also boost the end-to-end acoustic-to-word CTC model to achieve much better WER than the traditional context-dependent phoneme CTC model decoded with a very large-sized language model.

% \vfill\pagebreak
\clearpage
\bibliographystyle{IEEEbib}
\balance
\bibliography{strings,refs}

\end{document}